\documentclass[final,5p,times,twocolumn]{elsarticle}

\usepackage{amssymb}
\usepackage{amsmath}
\usepackage{multirow}
\usepackage{array}
\usepackage{booktabs}
\usepackage{amsmath,amsfonts}
\usepackage{pifont}
\usepackage{makecell}
\usepackage{graphicx}
\usepackage{wrapfig}
\usepackage{multicol}

\journal{Neurocomputing}
\date{November 3, 2024}

\begin{document}

\begin{frontmatter}



\title{SwinLip: An Efficient Visual Speech Encoder for Lip Reading Using Swin Transformer}

\author[label1]{Young-Hu Park}
\author[label2]{Rae-Hong Park}
\author[label1,label2]{Hyung-Min Park}
\affiliation[label1]{organization={Department of Artificial Intelligence, Sogang University},
            city={Seoul},
            postcode={04107},
            country={South Korea}}
\affiliation[label2]{organization={Department of Electronic Engineering, Sogang University},
            city={Seoul},
            postcode={04107},
            country={South Korea}}

\begin{abstract}
This paper presents an efficient visual speech encoder for lip reading. While most recent lip reading studies have been based on the ResNet architecture and have achieved significant success, they are not sufficiently suitable for efficiently capturing lip reading features due to high computational complexity in modeling spatio-temporal information. Additionally, using a complex visual model not only increases the complexity of lip reading models but also induces delays in the overall network for multi-modal studies (e.g., audio-visual speech recognition, speech enhancement, and speech separation). To overcome the limitations of Convolutional Neural Network (CNN)-based models, we apply the hierarchical structure and window self-attention of the Swin Transformer to lip reading. We configure a new lightweight scale of the Swin Transformer suitable for processing lip reading data and present the SwinLip visual speech encoder, which efficiently reduces computational load by integrating modified Convolution-augmented Transformer (Conformer) temporal embeddings with conventional spatial embeddings in the hierarchical structure. Through extensive experiments, we have validated that our SwinLip successfully improves the performance and inference speed of the lip reading network when applied to various backbones for word and sentence recognition, reducing computational load. In particular, our SwinLip demonstrated robust performance in both English LRW and Mandarin LRW-1000 datasets and achieved state-of-the-art performance on the Mandarin LRW-1000 dataset with less computation compared to the existing state-of-the-art model.
\end{abstract}



\begin{keyword}


Lip reading, visual speech recognition, visual speech encoder, Swin Transformer, ResNet.
\end{keyword}

\end{frontmatter}



\section{Introduction}
Lip reading, also known as visual speech recognition (VSR), recognizes a speaking message from the movement of the lips without audio information. This technology is becoming increasingly important in the field of speech-related technology because it can be a very useful means for robust speech recognition in noisy environments or communication with deaf or hearing-impaired people~\cite{lay2008application,woodhouse2009review}. 
Additionally, it can be combined with the acoustic model of the speech recognition system to improve recognition performance in noisy environments~\cite{afouras2018deep,ma2021end}, and lip reading has recently shown strong performance, being applied to various applications such as speech enhancement and speech separation~\cite{michelsanti2021overview, zhu2023real, lee2024seeing}.

Lip reading requires recognizing speech with only visual information without audio information and is a very challenging task especially due to homophonic ambiguity. Recent advances in deep learning techniques have solved these problems to some extent, enabling deep neural networks (DNNs) to extract lip reading video features. Therefore, various studies have been proposed to extend convolutional-neural-network (CNN)-based models to lip reading~\cite{chung2017lip,afouras2018dlr, stafylakis2017combining, martinez2020lipreading}, which outperformed hand-crafted methods. In particular, the representative model of CNN, 2D ResNet, has been widely used in lip reading, which is an application of a 2D image recognition model to high-dimensional video recognition tasks. 

These approaches commonly involve extracting visual features from raw video, where CNN operations are repeated for each frame along the time axis, leading to a significant increase in computational load due to the time dimension.
However, lip reading requires efficient computation because it needs to facilitate real-time interaction for practical applications. Furthermore, it is helpful to exploit not only the shape of the mouth but also the global movements of the surrounding areas (e.g., jaw, nose, and cheeks) when performing lip reading. However, the operation method of CNN's convolutional filters places more emphasis on local information, which can result in the loss of important global information during this process. Therefore, there is a need to develop a visual encoder that can effectively extract visual features with low computational cost.

There have been studies introducing CNN-based models like ShuffleNet~\cite{ma2021towards} and EfficientNet~\cite{koumparoulis2022accurate} for lip reading as lightweight models. However, ShuffleNet was reported to have relatively poor performance on lip reading, and the performance of EfficientNet was only reported when integrated with a Transformer decoder, making it difficult to independently verify 
the performance of the model as a visual encoder for lip reading. Additionally, while the Convolutional vision Transformer (CvT)~\cite{wu2021cvt} from the Vision Transformer (ViT) family was applied to lip reading in an attempt to capture the global information, it showed similar performance to CNN-based lip reading methods and lacked efforts to reduce computational burden. Other methods~\cite{hao2021use, tian2022lipreading, kim2021cromm, kim2022distinguishing} have focused solely on improving recognition accuracies by integrating additional modules with lip reading models or indirectly using audio information. 

In this paper, we aim to develop an efficient visual speech encoder for lip reading that can both reduce computational load and improve recognition performance.
We propose a new visual encoder called SwinLip for lip reading by applying the Swin Transformer~\cite{liu2021swin} structure.
The Swin Transformer uses a shifted window attention mechanism to perform attention independently within windows, achieving a first-order computational complexity model. The Swin Transformer introduces a hierarchical structure composed of a total of four stages, which is optimized for the image size of 224$\times$224 pixels in the ImageNet dataset~\cite{deng2009imagenet}. Since lip reading datasets only need to include the area around the face, they are mostly composed of 96$\times$96 pixels of frame image sequences. Thus, we modify the window and patch sizes and the hierarchical structure of the Swin Transformer to design a structure suitable for processing small-size lip reading images.

Additionally, since lip reading is spatio-temporal data, inputting it directly into a 2D Swin Transformer would not allow the model to consider temporal information. Therefore, we introduce a 3D Spatio-Temporal Embedding Module, composed of a single 3D CNN layer, at the front of the model to add temporal embeddings to the data. In this process, we do not squeeze the input image and keep the shape intact for smooth patch operation of the Swin Transformer. Moreover, to bolster our visual encoder's ability to grasp the temporal aspects of speech, we integrate a 1D Convolutional Attention Module at the final layer of the Swin Transformer. Furthermore, to facilitate streaming operations for the VSR and audio-visual speech recognition (AVSR) models, we develop models capable of causal operations by eliminating self-attention and batch normalization (BN).

To demonstrate the capability of our model as the feature extraction encoder universally applicable to a wide range of tasks, we tested its performance by integrating it with various backend decoders. We conducted experiments by combining the SwinLip encoder with decoders~\cite{ma2021lip, feng2020learn, burchi2023audio} to evaluate the performance on English and Chinese lip reading datasets, including LRW~\cite{chung2017lip}, LRW-1000~\cite{yang2019lrw}, and the sentence-level English datasets of LRS2~\cite{chung2017blip} and LRS3~\cite{afouras2018lrs3}. In all cases, our SwinLip reduced FLoating point OPerations (FLOPs) and improved recognition performance. Notably, our SwinLip, using only visual information, achieved better performance on the LRW-1000 (Mandarin) dataset with less computational cost than the previous models. Additionally, we compared the performance of SwinLip with various vision backbones applied to lip reading. The computational cost and recognition performance of recently proposed CNN-based model~\cite{xiao2020deformation}, MLP-based models~\cite{tolstikhin2021mlp, chen2022cyclemlp}, and ViT-based models~\cite{xiao2020deformation} used as a visual frontend for lip reading were not as effective as our SwinLip, both in terms of recognition accuracy and computational cost. This demonstrated that our proposed SwinLip is an efficient model suitable for lip reading and can be applied and utilized in various lip-reading studies.

This paper makes the following contributions:
\begin{itemize}
\item We build SwinLip, a novel lip reading visual speech encoder based on Swin Transformer. This is the first time that a Swin Transformer architecture has been introduced for lip reading.
\item We introduce a novel configured 3D Spatio-Temporal Embedding Module for processing lip reading videos in the ViT and a 1D Convolutional Attention Module capable of capturing the temporal characteristics of utterances.
\item Our SwinLip model is easily used as the visual frontend for various backbones for word and sentence recognition, achieving improved recognition performance with less computational complexity.
\item The proposed SwinLip model achieved a new state-of-the-art performance on the LRW-1000 dataset while reducing the computational load with a comparable number of model parameters compared to the previous state-of-the-art model.
\end{itemize}

\section{Related Works}
\subsection{Lip Reading}
Traditionally, the lip reading systems extracted hand-crafted or well-known image features such as the discrete cosine transform (DCT) to encode the shape of lips and then fed them to a decoder such as the hidden Markov model (HMM) for temporal information modeling~\cite{puviarasan2011lip,potamianos2003recent,hong2006pca,fernandez2018survey}.
With the development of computing resources and deep-learning technologies, lip reading methods have also been replaced by deep-learning-based models. Feature extraction such as the DCT was replaced by CNNs, such as VGG~\cite{chung2017lip}, and the HMM was replaced by the Long Short-Term Memory (LSTM)~\cite{wand2016lipreading,petridis2017end}. 
Like other fields, end-to-end deep-learning models for lip reading have been developed so that all learning variables are optimized by minimizing cost functions for lip reading.
In particular, 3D CNN was mainly used to effectively encode spatio-temporal information corresponding to pronunciation sequences from visual speech data and combined with a Transformer architecture~\cite{afouras2018dlr}.  

Recently, many methods were developed based on a network combining shallow 3D CNN layers and a 2D CNN block~\cite{chung2017lip,stafylakis2017combining}. In spatial modeling using 2D ResNet, features of lip reading data were extracted and fed to Bidirectional Gated Recurrent Units (Bi-GRUs)~\cite{xu2018lcanet,zhang2020can} or various models of Temporal Convolutional Networks (TCNs)~\cite{xiao2020deformation,martinez2020lipreading,ma2021towards,ma2022training,wang2022lip} to follow a time modeling process for temporal modeling in VSR.

On the other hand,~\cite{hao2021use} inserted the Temporal Shift Module (TSM)~\cite{lin2019tsm} into the ResNet to extend the temporal receptive field. To improve spatial modeling,~\cite{tian2022lipreading} proposed collaborative learning to divide the overall feature space according to its spatial location and then assign weights to local features based on preference between local and global features. 
\cite{kim2022distinguishing, yeo2023multi} used audio information additionally, unlike previous studies that used only visual information. They proposed an audio-visual lip reading structure based on the memory network using the audio encoding vector as the value memory of the attention mechanism.
\cite{ma2022training} showed that temporal masking based on SpecAugment~\cite{park2019specaugment} was effective for lip reading and used word boundary indicators to configure the temporal model. 
Although the combination of 3D CNN and ResNet modeled spatio-temporal information more effectively, it was not efficient in terms of computational complexity. Lip reading, unlike audio speech recognition, requires the processing of high-dimensional image sequence inputs with spatio-temporal complexity. This makes it more challenging to train a large-scale end-to-end model for lip reading or AVSR due to computational loads.

\subsection{Vision-Transformer-like Architectures}
To solve the problem of long-term dependency of sentences in the natural language processing field,~\cite{vaswani2017attention} introduced an attention-based encoder-decoder structure called Transformer~\cite{bahdanau2014neural}. The Transformer architecture has successfully replaced Recurrent Neural Networks (RNNs) and LSTMs, achieving outstanding performance in the speech recognition field. In order to apply the Transformer model to computer vision tasks,~\cite{dosovitskiyimage} presented the ViT in which images are partitioned into patch units, tokenized, and fed to the self-attention block of the Transformer model. After the success of the ViT, there have been several studies to replace CNNs with a self-attention-based architecture.

The Data-efficient image Transformer (DeiT)~\cite{hugo2021training} introduced a knowledge distillation (KD) learning method to improve the ViT that required pre-training with vast amounts of data. Influenced by CNN's multi-scale resolution learning method, the Pyramid Vision Transformer (PVT)~\cite{wang2021pyramid} proposed spatial-reduction attention that reduced computational complexity by constructing a Transformer network inspired by a pyramid structure and reducing sequences at each stage. In addition, the CvT~\cite{wu2021cvt} improved feature learning capabilities at various levels by introducing convolution-based token embedding and projection methods to add hierarchical effects of CNN to ViT. The Swin Transformer~\cite{liu2021swin} introduced a hierarchical structure to address the quadratic computational complexity problem of ViT and proposed shifted window attention, inspired by the sliding kernel method of CNNs. This approach achieved linear computational complexity across the entire network and facilitated extension to various vision tasks.

Recently, the Deformable Attention Transformer (DAT)~\cite{xia2022vision} proposed a deformable attention module for modeling local and global relationships. DAT could select the location of key and value pairs by offset networks to perform attention operations in a data-dependent manner. However, the effect of deformable attention could be limited to downsizing vision tasks that required modeling fine local information such as lip reading, and additional operations were required in offset networks.

Conformer~\cite{gulati2020conformer} was introduced in the lip reading field~\cite{ma2021end,burchi2023audio}. Conformer can effectively model the local and global dependencies of speech by adding convolution modules to the self-attention-based Transformer structure. 
In this paper, inspired by this, we propose to introduce a 1D Convolutional Attention Module into the Swin Transformer lip reading encoder to enhance its ability to extract features from data that include a temporal dimension. Our proposed method is evaluated in combination with various lip reading backends.

\begin{figure*}[!t]
    \centerline{\includegraphics[width=1.0\linewidth]{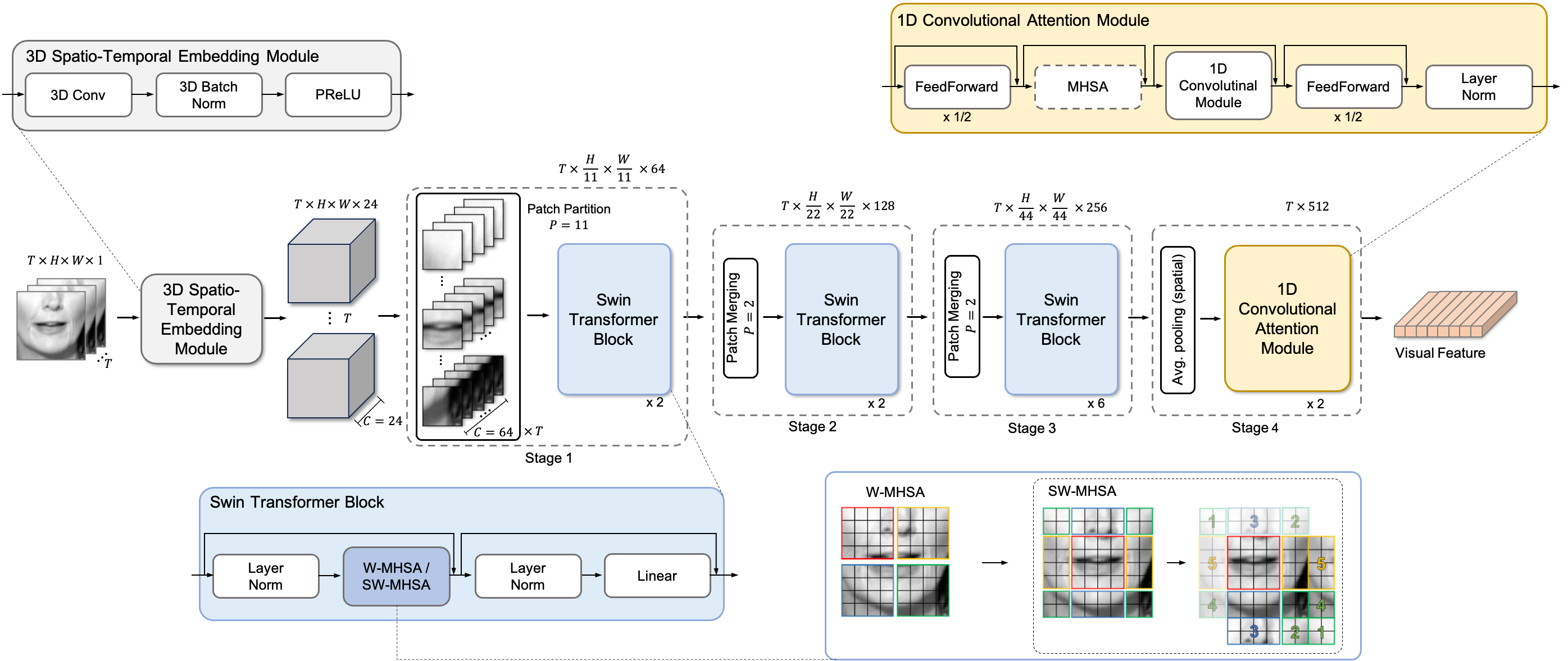}}
    \caption{The overall structure of the proposed SwinLip model for lip reading. The model consists of a combination of a 3D Spatio-Temporal Embedded Module, Swin Transformer and a 1D Convolutional Attention Module. $P$ is the patch size. In the streaming mode, the MHSA layer is removed from the 1D Convolutional Attention Module.}
    \label{figure_1} 
\end{figure*}

\section{Proposed Method}
The structure of the proposed SwinLip architecture is shown in Figure~\ref{figure_1}. Our SwinLip first maintains the shape of the lip reading data through a 3D Spatio-Temporal Embedding Module and performs embedding only along the channel axis. This corresponds to process the temporal information in the 2D Swin Transformer. In addition, we modify the Swin Transformer structure to handle small-sized lip reading data. The hierarchical structure of the Swin Transformer reduces the size of the image by half at each stage. If we apply four stages to lip reading data as in the conventional Swin Transformer, the size of the input image is sufficiently reduced by the third stage, causing redundant calculations in the last stage. Therefore, we replace the last stage of the Swin Transformer with a 1D Convolutional Attention Module to enhance the temporal feature extraction capability of the model. Furthermore, we present a streaming model by removing Multi-Head Self-Attention (MHSA) and BN from the 1D Convolutional Attention Module.

\begin{figure}[!t]
    \centerline{\includegraphics[scale=0.55]{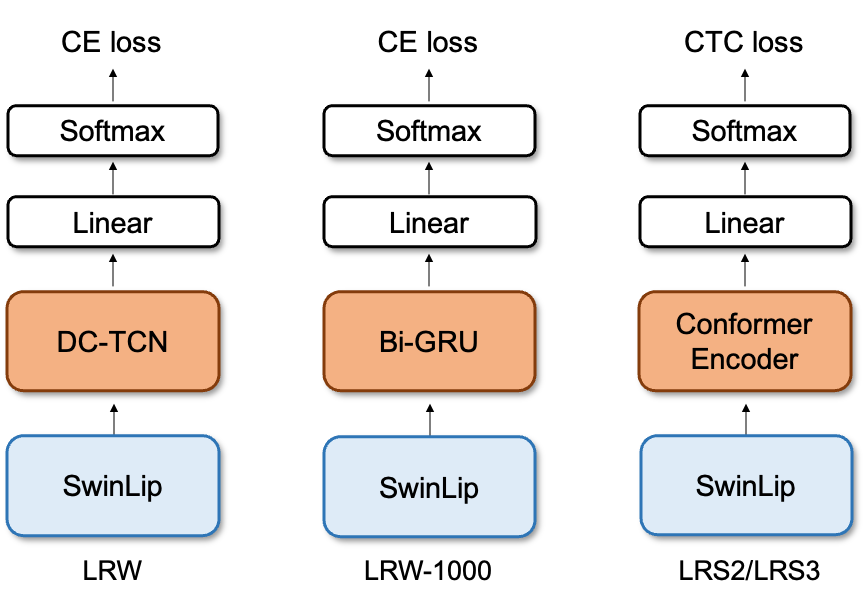}}
    \caption{Backend structures combined with the proposed SwinLip. They are temporal backend structures for lip reading of English words (LRW), Mandarin words (LRW-1000), and English sentences (LRS2/LRS3), respectively.}
    \label{figure_2} 
\end{figure}

We connect three backend decoders, DC-TCN, Bi-GRU, and Conformer, which achieved good performance in English word, Chinese word, and English sentence recognition, respectively, to the SwinLip. Each backend decoder is shown in Figure~\ref{figure_2}. Each recognition model is constructed by connecting these with the SwinLip architecture. The detailed configuration of the SwinLip architecture is summarized in Table~\ref{tbl1}.

\subsection{3D Spatio-Temporal Embedding Module} 
In most lip reading studies~\cite{martinez2020lipreading, ma2021towards, ma2022training, huang2022novel}, to input lip reading video data into a 2D ResNet, the 3D kernel size of the 3D CNN layer was $(5,7,7)$, with stride sizes of $(1,2,2)$, followed by BN and Rectified Linear Unit (ReLU) activation function. 
This process compresses the image before being input into the ResNet, which is necessary to prevent a significant increase in the computational load for the ResNet. However, such image compression limits the areas that can be observed in the ResNet, thereby hindering the effective extraction of visual information.

We introduce a 3D Spatio-Temporal Embedding Module to efficiently extract spatio-temporal information from lip reading video data.
Inspired by~\cite{koumparoulis2022accurate} on the 3D frontend for lip reading, we adjust the kernel size to $(3,5,5)$ to reduce information loss due to early downsampling of data. Additionally, we maintain the same tensor size as the input size $H \times W$ by setting the stride size to $(1,1,1)$ for the hierarchical structure and the patch-wise operation of the Swin Transformer, which is then fed to BN and the Parametric ReLU (PReLU) activation function~\cite{he2015delving}. This enables the Swin Transformer to view the entire lip reading image and effectively understand the global information. As shown in Figure~\ref{figure_1}, a $T$-frame input grayscale image sequence $X = \{x_1,x_2,...,x_T\}\in \mathbb{R}^{T \times H \times W \times 1}$ is embedded to visual feature map sequence $F_{\rm x}\in \mathbb{R}^{T \times H \times W \times 24}$ by the 3D Spatio-Temporal Embedding Module.

\begin{table}[t]
\centering
\footnotesize
\addtolength{\tabcolsep}{-4pt}
\renewcommand\arraystretch{1.5}
\begin{tabular}{c|c|c}
\Xhline{3\arrayrulewidth}
\multicolumn{2}{c|}{Network}        & SwinLip \\ \hline\hline
\multicolumn{2}{c|}{\begin{tabular}[c]{@{}c@{}}3D Spatio-Temporal \\ Embedding Module\end{tabular}} &
  \begin{tabular}[c]{@{}c@{}} Conv3D~(Kernel: $3\times5\times$5, \\Stride: $1\times1\times$1, $C: 24$), $BN$, $PReLU$\end{tabular} \\ \hline
\multirow{7}{*}{\begin{tabular}[c]{@{}c@{}}Swin \\ Transformer\end{tabular}} &
  Patch Partition & \begin{tabular}[c]{@{}c@{}} $C: 64$, $P: 11$ \end{tabular} \\ \cline{3-3}
\multicolumn{1}{c|}{}         & Stage 1        & \begin{tabular}[c]{@{}c@{}} $C: 64$, $P: 2$, $N: 2$ \\ Window: 4, Head: 2\end{tabular}   \\ \cline{3-3}
\multicolumn{1}{c|}{}         & Stage 2        & \begin{tabular}[c]{@{}c@{}} $C: 128$, $P: 2$, $N: 2$ \\ Window: 4, Head: 4\end{tabular} \\ \cline{3-3}
\multicolumn{1}{c|}{}         & Stage 3        & \begin{tabular}[c]{@{}c@{}} $C: 256$, $P: 2$, $N: 6$ \\ Window: 2, Head: 8\end{tabular} \\ \hline
\multicolumn{1}{c|}{\begin{tabular}[c]{@{}c@{}}1D Convolutional \\ Attention Module\end{tabular}}& Stage 4        & \begin{tabular}[c]{@{}c@{}} $C: 512$, $N: 2$, Head: 16\end{tabular} \\ \hline\hline
\multicolumn{2}{c|}{FLOPs (G) / Parameters (M)}  & 1.92 / 12.46      \\   \Xhline{3\arrayrulewidth}
\end{tabular}
\caption{Details of our SwinLip architecture. $C$: the embedding dimension of each network, $P$: the patch size, $N$: the number of layers. FLOPs and parameters are counted only for the visual encoder network.}\label{tbl1}
\end{table}

\subsection{Swin Transformer for Lip Reading}
To effectively capture and encode the global information of continuous lip movements, we introduce the Swin Transformer to lip reading. Since the Swin Transformer is optimized for images of size $224 \times 224$, it cannot be directly applied to lip reading data which has a pre-processed resolution of $88 \times 88$. Therefore, we adjust the initial patch size to $11 \times 11$. By adopting a large patch size, the model can understand lip movements more globally. Furthermore, since this method reduces the total number of patches that the encoder needs to process, it is sufficient to extract features by processing up to only the third stage of the hierarchical structure of the conventional Swin Transformer. Therefore, we remove the last layer of the Swin Transformer because only operations for expanding the hidden dimension are performed in the last layer, which may result in performance degradation.

As the stage progresses, the patches are merged by concatenating each group of neighboring patches of size $2 \times 2$. By the patch partition, the number of patches obtained from a frame image is $(\frac{H}{11}, \frac{W}{11}):=(h,w)$. After $24$-channel patches are projected to the $64$ channel dimension by a 2D CNN, we get $z_{f}\in \mathbb{R}^{T \times h \times w \times 64}$. In addition, shifted-window attention is performed for fine lip movements by reducing the window size. We modify the window size to $M \times M$ to obtain local attention features that capture fine lip movements. For the initial window size, we use $M = 4$. The flow of the proposed model is shown in Figure~\ref{figure_1}.
 
In each Swin Transformer block, there exist learnable weight matrices $W_{q}, W_{k}, W_{v}\in \mathbb{R}^{C \times C}$ to project local window features $z^\prime_{t,s}\in \mathbb{R}^{M^{2} \times C}$, $t \in \{1,..., T\}$, $s \in S = \{1,...,{hw}{M^2}\}$ as query ($Q_{z^\prime}$), key ($K_{z^\prime}$), and value ($V_{z^\prime}$), respectively, where $T$ is the total input frames, $C$ is the hidden dimension channels, and ${hw}/{M^2}$ is the total number of windows. The attention mechanism is calculated as follows:
\begin{align}\label{att}
&\text{Attention} = \text{Softmax}(Q_{z^\prime}K^T_{z^\prime} / \sqrt{d} + B)V_{z^\prime},
\end{align}
where $d = C / Heads$ and $B \in \mathbb{R}^{M^{2} \times M^{2}}$ is a relative position bias used in~\cite{liu2021swin}. This self-attention mechanism is repeated as many times as the number of heads and the results are concatenated to construct an output of MHSA. 

In addition, window partitioning is applied to complement the connectivity between independent local window attention. Like the Swin Transformer Block in Figure~\ref{figure_1}, the standard local window attention is applied in the first layer. Then, the window moves the pixel position to $(\lfloor\frac{M}{2}\rfloor, \lfloor\frac{M}{2}\rfloor)$ in the next layer to construct a shifted-window. The number of parts that are not included in these post-move windows is adjusted to be equal to the total number of windows in the standard window attention mechanism by using the cyclic shift method.
The Swin Transformer block composed of this shifted-window approach is calculated as follows:
\begin{align}\label{swin}
&\hat{z}_{t, S, l} = \text{W-MHSA}(\text{LN}(z^\prime_{t, S, l-1})) + z^\prime_{t, S, l-1},  \\
&z^\prime_{t, S, l} = \text{MLP}(\text{LN}(\hat{z}_{t, S, l}))+\hat{z}_{t, S, l},   \\
&\hat{z}_{t, S, l+1} = \text{SW-MHSA}(\text{LN}(z^\prime_{t, S, l})) + z^\prime_{t, S, l},  \\
&z^\prime_{t, S, l+1} = \text{MLP}(\text{LN}(\hat{z}_{t, S, l+1}))+\hat{z}_{t, S, l+1}, 
\end{align}
where $z^\prime_{t, S, l-1}\in\mathbb{R}^{\frac{hw}{M^2} \times M^{2} \times C}$ is the output feature for layer $l-1$ at the $t$-th frame. Specifically, $\hat{z}_{t, S, l}$ and $\hat{z}_{t, S, l+1}$ are the outputs of the Window(W)-MHSA and Shifted-Window(SW)-MHSA with a Layer Norm (LN) of $z^\prime_{t, S, l-1}$ and $z^\prime_{t, S, l}$, respectively, with skip connection. 
Also, $z^\prime_{t, S, l}$ and $z^\prime_{t, S, l+1}$ are the outputs of the Multi-Layer Perceptron (MLP) module with an LN of $\hat{z}_{t, S, l}$ and $\hat{z}_{t, S, l+1}$, respectively, with skip connection.
The detailed configuration of the proposed SwinLip is shown in Table~\ref{tbl1}.

\subsection{1D Convolutional Attention Module}
The output of the Swin Transformer is a set of feature vectors over time. Since this output extracts features of images independent of the time axis, it cannot capture continuous lip movements according to speech. Therefore, similar to studies~\cite{wang2018non,yin2020disentangled} that integrate non-local blocks into the original architecture, we introduce a 1D Convolutional Attention Module at the last stage removed from the Swin Transformer. This enables the model to encode the feature vectors along the time axis, thus understanding information according to an utterance. The structure of the 1D Convolutional Attention Module is inspired by the Conformer~\cite {gulati2020conformer} which has shown impressive performance in the speech recognition field.

The Swin Transformer output feature $g_z \in \mathbb{R}^{T \times 512}$ after average pooling is fed to the 1D Convolutional Attention Module, and the output feature has the same dimension as the input $g_z$. In the 1D Convolutional Attention Module, a Feed-Forward Network (FFN), an MHSA module, a convolution module, and an FFN are stacked in order with skip connection, followed by an LN. The 1D Convolutional Attention Module can be expressed as
\begin{align}\label{swin}
&y^\prime = g_z +\frac{1}{2}\text{FFN}(g_z),  \\
&y^{\prime\prime} = y^\prime + \text{MHSA}(y^\prime),   \\
&y^{\prime\prime\prime} = y^{\prime\prime} + \text{CNN}(y^{\prime\prime}),  \\
&y = \text{LN}(y^{\prime\prime\prime} + \frac{1}{2}\text{FFN}(y^{\prime\prime\prime})) \in \mathbb{R}^{T\times 512}, 
\end{align}
where $y^\prime$, $y^{\prime\prime}$, $y^{\prime\prime\prime} \in \mathbb{R}^{T \times 512}$, with all intermediate outputs kept at 512 dimensions to minimize computational complexity, and CNN denotes the 1D Convolutional Module, as illustrated in Figure 1. The input feature $g_z$, after average pooling, is processed by each module to extract temporal features along the time axis. The number of attention heads in the MHSA module is set to $16$ according to the hierarchical structure setting of the Swin Transformer.

Furthermore, the streaming model of SwinLip can be obtained by removing the MHSA and BN phase in the CNN from the 1D Convolutional Attention Module of the previously described model.

\section{Experiments}
\subsection{Datasets}
We have conducted experiments on four large-scale lip reading datasets from word to sentence levels. The LRW~\cite{chung2017lip} and LRW-1000~\cite{yang2019lrw} datasets were experimented for words while the LRS2~\cite{chung2017blip} and LRS3~\cite{afouras2018lrs3} datasets were considered for sentences.

The LRW is a word-level English lip reading benchmark dataset uttered by more than 1,000 speakers, which was collected from BBC TV news programs and other sources. It consists of up to 1,000 utterances of 500 different words. All video clips are 1.16 seconds long corresponding to 29 frames. For each video clip, there is a label that contains information on the speech activity interval. The LRW-1000 is a word-level Mandarin lip reading benchmark dataset. It consists of 718,018 video clips of 1,000-class word utterances by more than 2,000 individual speakers. Each video clip consists of 40 frames, and the data were collected in a similar way to the LRW. The LRS2 and LRS3 are sentence-level English lip reading benchmark datasets. The LRS2 is 224 hours of speech videos collected from BBC TV while the LRS3 is 438 hours of speech videos collected from TED and TEDx on YouTube. Each video clip contains speech sentences of various lengths, and the same face landmark extraction and cropping method as in the LRW was applied to the clips. 

\subsection{Data Pre-processing} 
In the same manner as in~\cite{ma2022training}, we detected faces in the video and extracted 68 landmarks. The detected landmarks were used to crop each frame of the video by a mouth Region Of Interest (ROI) of a $96 \times 96$-pixel bounding box. Then, each cropped image was converted from RGB to gray level.

When the model was trained, input images were randomly cropped to $88 \times 88$ pixels with normalization and horizontally flipped with the probability value of 0.5 for data augmentation. Input images for validation were cropped at the center region of $88 \times 88$ pixels with normalization. Other pre-processing steps followed those used in the baselines for the corresponding datasets.

\subsection{Training Details} 
We combined our SwinLip with three temporal backends, DC-TCN~\cite{ma2022training}, Bi-GRU~\cite{feng2020learn}, and Conformer encoder~\cite{burchi2023audio}, and trained the models for the LRW, LRW-1000, and LRS2/LRS3 datasets, respectively. The Cross-Entropy (CE) loss was used for word-level training while the Connectionist Temporal Classification (CTC) loss was used for sentence-level training (see Figure~\ref{figure_2}). 
The models for word recognition were trained for 100 epochs with a batch size of 32 using AdamW optimizer~\cite{loshchilov2018decoupled} with a weight decay of 1e-2. The learning rate was linearly increased during the first warm-up stage from the initial value of 2e-7 and decreased with a cosine annealing strategy~\cite{loshchilov2017sgdr}. We set the warm-up epochs to 8 and 12 for the LRW and LRW-1000 datasets, respectively, and the peak learning rate to 3e-4.
For the sentence-level model training, we pre-trained our SwinLip for 10 epochs on the LRW dataset and then trained the whole model using all of the pre-training, training, and validation sets from the LRS2 and LRS3 datasets. The trained model was evaluated for each of the LRS2 and LRS3 datasets. The visual-only (VO) and audio-visual (AV) models were trained for 105 and 80 epochs, respectively, with a batch size of 8 using AdamW optimizer with a weight decay of 1e-2, $\beta_1 = 0.9$, and $\beta_2 = 0.98$. We used the Noam scheduler~\cite{vaswani2017attention} with the warm-up steps to 10k and the peak learning rate to 1e-3.
In the AV mode, the audio frontend transformed raw audio waveforms to mel-spectrograms using a short-time Fourier transform and fed them to the 2D CNN. The audio and video features were concatenated and fed to a fusion module composed of linear layers and Swish activation functions~\cite{ramachandran2017searching}.
We used the additional language model for the decoding step as in~\cite{burchi2023audio}. Also, Stochastic Weight Mean~\cite{izmailov2018averaging} was used to average the weights for the last 5 and 10 epochs in the VO and AV models, respectively.

\section{Results}
In this section, we conducted experiments to directly compare the performance of the proposed SwinLip with model architectures commonly used in lip reading studies~\cite{feng2020learn, burchi2023audio, ma2022training}. Studies that used supplementary audio information~\cite{kim2022distinguishing} or KD~\cite{ma2022training} were excluded from our experiments to ensure a fair comparison based solely on visual information.

\subsection{Effects of SwinLip}
In this experiment, we demonstrated that replacing the visual encoders of existing VSR models with our proposed SwinLip could improve recognition performance with less computational load. Tables~\ref{tbl2}, \ref{tbl2-1}, and \ref{tbl2-2} compared the presented SwinLip visual encoder with ResNet18-based models which had been successfully used in lip reading. The performance was evaluated in terms of word accuracy by using three different backend structures for the tasks of word-level English, word-level Mandarin, and sentence-level English lip reading as indicated in Figure~\ref{figure_2}. 

\begin{table}[t]
\centering
\small
\addtolength{\tabcolsep}{-2.pt}
\renewcommand\arraystretch{1.2}
\begin{tabular}{cccccc}
\Xhline{3\arrayrulewidth}
\multicolumn{6}{c}{\textbf{LRW}}                                 \\ \hline
\multicolumn{1}{c}{\textbf{Method}} & \multicolumn{2}{c}{\begin{tabular}[c]{@{}c@{}}\textbf{Visual}\\\textbf{Encoder}\end{tabular}} & \multicolumn{2}{c|}{\textbf{Acc. (\%)}} & \textbf{WB} \\ \hline\hline
\multicolumn{1}{c}{\begin{tabular}[c]{@{}c@{}}Baseline~\cite{ma2022training}\end{tabular}} & \multicolumn{2}{c}{ResNet18} & \multicolumn{2}{c|}{89.52}& \multirow{3}{*}{\ding{55}}   \\ 
\multicolumn{1}{c}{\textbf{Ours}} & \multicolumn{2}{c}{\textbf{SwinLip}} &\multicolumn{2}{c|}{\textbf{90.67}}& \\
\multicolumn{1}{c}{\textbf{Ours}} & \multicolumn{2}{c}{\begin{tabular}[c]{@{}c@{}}\textbf{SwinLip-Streaming}\end{tabular}} &\multicolumn{2}{c|}{\textbf{90.30}}& \\ \hline
\multicolumn{1}{c}{Baseline~\cite{ma2022training}} & \multicolumn{2}{c}{ResNet18} & \multicolumn{2}{c|}{91.65} & \multirow{2}{*}{\ding{51}} \\
\multicolumn{1}{c}{\textbf{Ours}} &\multicolumn{2}{c}{\textbf{SwinLip}} & \multicolumn{2}{c|}{\textbf{92.43}}&    \\ \Xhline{3\arrayrulewidth}
\end{tabular}
\caption{Word accuracies on tasks of word-level English lip reading (LRW) for the presented SwinLip and ResNet18-based visual encoders. WB represents the models that used word boundary information.
}\label{tbl2}
\end{table}

\begin{table}[t]
\centering
\small
\addtolength{\tabcolsep}{2.pt}
\renewcommand\arraystretch{1.2}
\begin{tabular}{cccccc}
\Xhline{3\arrayrulewidth}
\multicolumn{6}{c}{\textbf{LRW-1000}}                            \\ \hline\hline
\multicolumn{1}{c}{\textbf{Method}} & \multicolumn{2}{c}{\begin{tabular}[c]{@{}c@{}}\textbf{Visual}\\\textbf{Encoder}\end{tabular}} & \multicolumn{2}{c|}{\textbf{Acc. (\%)}} & \textbf{WB} \\ \hline\hline
\multicolumn{1}{c}{\begin{tabular}[c]{@{}c@{}}Baseline~\cite{feng2020learn}\end{tabular}} & \multicolumn{2}{c}{ResNet18} & \multicolumn{2}{c|}{45.70}  & \multicolumn{1}{c}{\multirow{2}{*}{\ding{55}}} \\ 
\multicolumn{1}{c}{\textbf{Ours}} & \multicolumn{2}{c}{\textbf{SwinLip}} & \multicolumn{2}{c|}{\textbf{48.09}}  & \\  \hline
\multicolumn{1}{c}{Baseline~\cite{feng2020learn}} & \multicolumn{2}{c}{ResNet18} & \multicolumn{2}{c|}{55.16}  & \multicolumn{1}{c}{\multirow{2}{*}{\ding{51}}} \\
\multicolumn{1}{c}{\textbf{Ours}} & \multicolumn{2}{c}{\textbf{SwinLip}} & \multicolumn{2}{c|}{\textbf{59.41}}  & \\ \Xhline{3\arrayrulewidth}
\end{tabular}
\caption{Word accuracies on tasks of word-level Mandarin lip reading (LRW-1000) for the presented SwinLip and ResNet18-based visual encoders. WB represents the models that used word boundary information.
}\label{tbl2-1}
\end{table}

\begin{table}[t!]
\centering
\small
\addtolength{\tabcolsep}{2.pt}
\renewcommand\arraystretch{1.2}
\begin{tabular}{cccccc}
\Xhline{3\arrayrulewidth}
\multicolumn{6}{c}{\textbf{LRS2/LRS3}}                              \\ \hline
\multicolumn{1}{c}{\textbf{Method}} & \multicolumn{2}{c}{\begin{tabular}[c]{@{}c@{}}\textbf{Visual}\\\textbf{Encoder}\end{tabular}}  & \multicolumn{2}{c|}{\begin{tabular}[c]{@{}c@{}}\textbf{WER (\%)}\\ \textbf{LRS2/LRS3}\end{tabular}} & \multicolumn{1}{c}{\textbf{Modality}}                      \\ \hline\hline
\multicolumn{1}{c}{\begin{tabular}[c]{@{}c@{}}Baseline~\cite{burchi2023audio}\end{tabular}} & \multicolumn{2}{c}{ResNet18} & \multicolumn{2}{c|}{37.10 / 49.39}  & \multicolumn{1}{c}{\multirow{2}{*}{VO}} \\ 
\multicolumn{1}{c}{\textbf{Ours}} & \multicolumn{2}{c}{\textbf{SwinLip}} & \multicolumn{2}{c|}{\textbf{37.01} / \textbf{48.88}}  & \\  \hline
\multicolumn{1}{c}{Baseline~\cite{burchi2023audio}} & \multicolumn{2}{c}{ResNet18} & \multicolumn{2}{c|}{~~3.06 / ~~2.73}  & \multicolumn{1}{c}{\multirow{2}{*}{AV}} \\ 
\multicolumn{1}{c}{\textbf{Ours}} & \multicolumn{2}{c}{\textbf{SwinLip}} & \multicolumn{2}{c|}{\textbf{~~2.86} / \textbf{~~2.37}}  & \\  \Xhline{3\arrayrulewidth}
\end{tabular}
\caption{Word Error Rates (WERs) on tasks of sentence-level English lip reading (LRS2/LRS3) for the presented SwinLip and ResNet18-based visual encoders. VO and AV denote visual-only and audio-visual modes, respectively.
}\label{tbl2-2}
\end{table}
In Table~\ref{tbl2}, our SwinLip architecture provided accuracy improvement of 1.15\% and 0.78\% over the baseline~\cite{ma2022training} based on ResNet18 which achieved the best performance on the LRW dataset, without and with a word boundary indicator, respectively. Furthermore, the streaming model with MHSA removed from the 1D Convolutional Attention Module also achieved an improvement of 0.78\% in accuracy over the baseline. This was a significant performance improvement on the LRW dataset and confirmed that our SwinLip visual encoder better captured the global information of lip movements.

For the LRW-1000 dataset, our model improved the accuracies by 2.39\% and 4.25\% compared to the baseline, without and with a word boundary indicator, respectively, as shown in Table~\ref{tbl2-1}. Mandarin lip reading is known to be more challenging than English lip reading as it contains various syllables with visually similar mouth shapes. Thus, these performance improvements demonstrated that our SwinLip could work well for languages other than English.

For the sentence-level lip reading using the LRS2 and LRS3 datasets, our SwinLip models achieved higher accuracies than the ResNet18-based models by 0.09\% and 0.51\% in VO modality and by 0.2\% and 0.36\% in AV modality, respectively, as shown in Table~\ref{tbl2-2}. This showed that our SwinLip was capable of effectively representing continuous speech beyond the word level. 

We observed that lip reading performance was consistently improved in all the experimental cases by replacing the ResNet18-based visual encoder with our SwinLip. This demonstrated that SwinLip, based on the Swin Transformer and capable of capturing global information efficiently, could successfully replace the existing visual encoder in lip reading and be compatible with different backend structures according to the tasks. The results also showed the benefits of using a strong visual encoder in the AVSR model where audio and video features were fused.

\subsection{Ablation Study}
\subsubsection{3D Spatio-Temporal Embedding Module}
We compared the computational load and recognition performance according to the kernel size of the 3D Conv layer used in our proposed 3D Spatio-Temporal Embedding Module. The experimental results are shown in Table~\ref{3dtem}. Most lip reading studies used a $(5,7,7)$ kernel size because it was necessary to reduce the image size with a stride larger than $1$ to decrease the computational load of ResNet. However, down-sampling the lip reading data substantially decreased the number of patches obtained through patch-wise operations, leading to convergence difficulties and performance degradation\footnote{Note that the word accuracy (89.52\%) of Baseline~\cite{ma2022training} in Table~\ref{tbl2} was the performance when the stride was 2.}. Therefore, we set the stride to 1 to maintain the shape of the image when inputting it into the model. Using our proposed $(3,5,5)$ kernel size, we observed a reduction in computational load by 0.92 GFLOPs and an improvement in performance by 0.71\%.

\subsubsection{1D Convolutional Attention Module}
We conducted an ablation study on our proposed 1D Convolutional Attention Module, and the results are shown in Table~\ref{tbl3}. When we removed the 1D Convolutional Attention Module from the SwinLip structure and used only the lightweight scale of the Swin Transformer, we observed a 1.29\% decrease in accuracy compared to the original SwinLip. Furthermore, when the 1D Convolutional Attention Module was not integrated within the Swin Transformer but applied as an external module at a subsequent stage, a 0.73\% decrease in accuracy was observed. This performance difference suggested that our proposed 1D Convolutional Attention Module, when combined with the last stage of the Swin Transformer, could create a significant synergy for lip reading. Moreover, this achieved higher performance than the baseline using ResNet, demonstrating that our configured SwinLip could serve as an efficient visual encoder for lip reading.

Our proposed 1D Convolutional Attention Module is based on the Conformer and includes MHSA operations, making it unsuitable for streaming operations. When we removed the MHSA operations to skip self-attention along the time axis for streaming operations, we observed a performance decrease of 0.37\%. Although there was a drop in recognition performance, it still surpassed the performance of baseline~\cite{ma2022training} commonly used in recent studies, proving that our SwinLip could be a powerful and efficient model even for streaming lip reading.

Additionally, while replacing the last layer of the Swin Transformer with the 1D Convolutional Attention Module, we removed the BN from the Convolutional Module in the Conformer. Since the Swin Transformer utilizes the LN technique, performing BN in the last layer of the model can make it difficult to maintain a consistent feature distribution, and excessive normalization may lead to increased computational load and performance degradation. When the BN was applied to the model, we observed that it reduced the computational efficiency of the SwinLip, resulting in a performance degradation of 0.14\%.

\begin{table}[!t]
\centering
\small
\addtolength{\tabcolsep}{-2.5pt}
\renewcommand\arraystretch{1.3}
\begin{tabular}{ccccc}
\Xhline{3\arrayrulewidth}
\multicolumn{1}{c}{\begin{tabular}[c]{@{}c@{}}\textbf{3D Conv} \textbf{Kernel Size}\end{tabular}}  & \textbf{FLOPs (G)}   & \textbf{Params (M)} &  \textbf{Acc. (\%)} \\  \hline\hline
    $(5, 7, 7)$             & ~2.84           & 12.47             & 89.96           \\ 
\textbf{$(3, 5, 5)$ (Ours)}      & ~\textbf{1.92}           & \textbf{12.46}            & \textbf{90.67}  \\ \Xhline{3\arrayrulewidth}
\end{tabular}
\caption{Performance comparison on the LRW dataset according to the 3D Conv kernel size of the 3D Spatio-Temporal Embedding Module used in the SwinLip encoder. Even for the different kernel size, the stride was identical at $(1, 1, 1)$. The kernel dimensions were represented in the form of $(T, H, W)$.}\label{3dtem}
\end{table}

\begin{table}[t]
\centering
\fontsize{7.2pt}{9.5pt}\selectfont 
\addtolength{\tabcolsep}{-5.5pt}
\renewcommand\arraystretch{1.4}
\begin{tabular}{lccc}
\Xhline{3\arrayrulewidth}
\multicolumn{1}{c}{\textbf{Method}}  & \textbf{FLOPs (G)}   & \textbf{Params (M)} &  \textbf{Acc. (\%)} \\    \hline\hline
\begin{tabular}[c]{@{}c@{}}Baseline~\cite{ma2022training}                      \end{tabular}  & 10.67         & 52.55          & 89.52          \\    \hline
\begin{tabular}[c]{@{}c@{}}\textbf{SwinLip}                                    \end{tabular}  & 3.40          & 53.84          & \textbf{90.67} \\ 
\begin{tabular}[c]{@{}c@{}} ~~(w/o) 1D Convolutional Attention Module          \end{tabular}  & 3.40          & 53.80          & 89.38          \\ 
\begin{tabular}[c]{@{}c@{}} ~~(w/) External-1D Convolutional Attention Module  \end{tabular}  & 3.48          & 56.98          & 89.94          \\ 
\begin{tabular}[c]{@{}c@{}} ~~(w/o) Multi-Head Self-Attention                  \end{tabular}  & \textbf{3.32} & \textbf{51.21} & 90.30          \\ 
\begin{tabular}[c]{@{}c@{}} ~~(w/) Batch Normalization                         \end{tabular}  & 3.40          & 53.85          & 90.53          \\    \Xhline{3\arrayrulewidth}
\end{tabular}
\caption{Ablation study of 1D Convolutional Attention Module in SwinLip in terms of word accuracies on the LRW dataset. }\label{tbl3}
\end{table}

\subsubsection{Inference Performance}
Table~\ref{tbl4} summarizes the FLOPs, number of parameters, and word accuracy of our SwinLip compared to a ResNet-based baseline~\cite{ma2022training}. Our SwinLip surpassed the performance of the baseline with approximately 1/3 fewer FLOPs and a similar number of parameters. Notably, our SwinLip showed significant competitiveness in terms of computational load. Moreover, our streaming model achieved higher performance with fewer FLOPs and parameters compared to the baseline. This demonstrated that our SwinLip was designed as a powerful model for lip reading with capturing global information efficiently.

\begin{figure}[!t]
    \centerline{\includegraphics[width=0.95\linewidth]{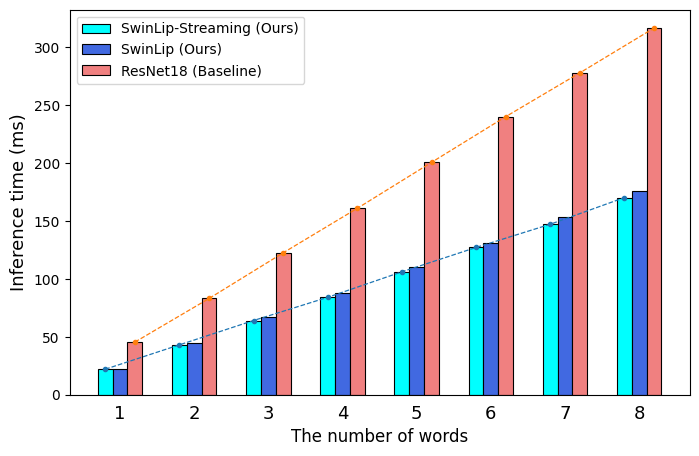}}
    \caption{Inference times of ResNet18-based and SwinLip visual speech encoders according to the number of words. Measurements were made on a system including a GPU of NVIDIA RTX 3090Ti.}
    \label{figure_3} 
\end{figure}

To evaluate the computational benefits in practical inference processes, Figure~\ref{figure_3} illustrates the inference time per word count in sentences for lip reading systems utilizing our proposed SwinLip and a ResNet18-based model. SwinLip and its streaming version showed significantly less inference times compared to the baseline. Particularly, ours increased much slowly in inference time as the number of words progressively increased, relative to the baseline. These findings demonstrated that our SwinLip could enhance the performance of real-world VSR systems while achieving faster inference. Since it is necessary to reduce the computational load for real-world applications, using our SwinLip can decrease the computational load of processing visual feature extraction.

\begin{table}[!t]
\centering
\small
\addtolength{\tabcolsep}{-4.pt}
\renewcommand\arraystretch{1.4}
\begin{tabular}{ccccc}
\Xhline{3\arrayrulewidth}
 \multicolumn{1}{c}{\begin{tabular}[c]{@{}c@{}}\textbf{Visual}\\ \textbf{Encoder}\end{tabular}}  & \begin{tabular}[c]{@{}c@{}}\textbf{FLOPs (G)}\\ {\scriptsize\textbf{Encoder (Total)}}\end{tabular}   & \begin{tabular}[c]{@{}c@{}}\textbf{Params (M)}\\ {\scriptsize\textbf{Encoder (Total)}}\end{tabular} &  \textbf{Acc. (\%)} \\ 
\hline\hline
ResNet18 (Baseline)   & ~9.2 (10.67)              & 11.18 (52.55)    & 89.52           \\  \hline
\textbf{SwinLip (Ours)}      & ~1.92 (3.40)   & 12.46 (53.84)             & \textbf{90.67}           \\ 
\begin{tabular}[c]{@{}c@{}} \textbf{SwinLip-Streaming (Ours)} \end{tabular}      & \textbf{~1.84 (3.32)}        & \textbf{9.82 (51.21)}             & 90.30  \\ \Xhline{3\arrayrulewidth}
\end{tabular}
\caption{Comparison of the FLOPs, numbers of parameters, and word accuracies of our SwinLip visual encoder and the ResNet18-based baseline on the LRW dataset. In common, DC-TCN was used as the backend.}\label{tbl4}
\end{table}

\subsection{Comparison with Other Vision Models}
\begin{figure}[t]
    \centerline{\includegraphics[width=1.\linewidth]{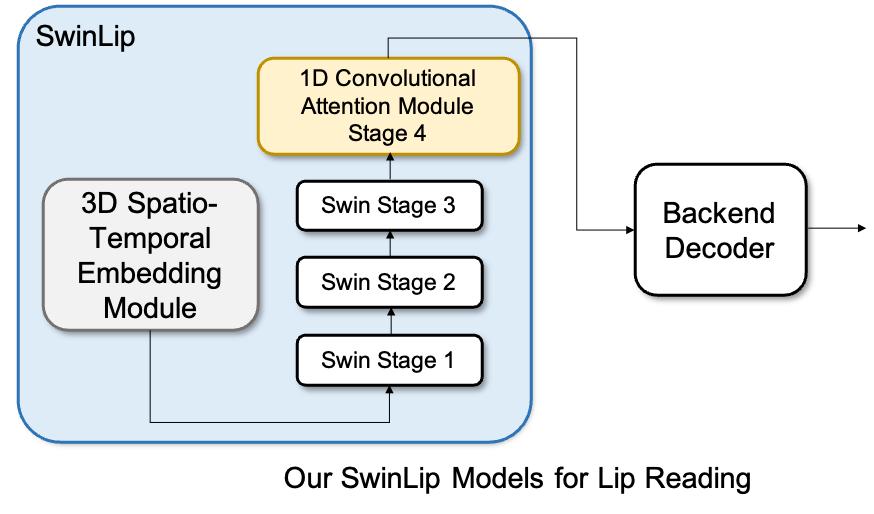}}
    \caption{An example of a visual encoder structure applying other vision models to lip reading compared to the SwinLip structure. To construct a model similar to SwinLip, we introduced the 3D Spatio-Temporal Embedded Model and the Conformer block.}
    \label{figure_4} 
\end{figure}

We have conducted experiments by directly applying various vision models that have been presented so far to lip reading, in order to demonstrate that our proposed SwinLip is the most suitable model for lip reading among multiple vision models. We selected models from the CNN, MLP, and Transformer families which are currently prominent in the vision field, applied them to lip reading, and evaluated their performance on the LRW dataset. For the CNN family, we used the ResNet18 baseline~\cite{ma2022training} along with the ConvNeXt~\cite{liu2022convnet}; for the MLP family, the MLP-Mixer~\cite{tolstikhin2021mlp} and Cycle-MLP~\cite{chen2022cyclemlp}; and for the Transformer family, the Swin Transformer~\cite{liu2021swin} and DAT~\cite{xia2022vision}. As shown in Figure~\ref{figure_4}, we added a 3D Spatio-Temporal Embedding Module at the frontend of these vision models to enable them to process lip reading video data. Additionally, we incorporated a Conformer block at the end of the models to add temporal modeling.

For the MLP-Mixer from the MLP family, we adopted the MLP-Mixer-B size to match the scale of the LRW dataset. Unlike the Swin Transformer, which is designed with a hierarchical structure, the MLP-Mixer maintains a fixed resolution across all layers without reducing the input image resolution and performs global interaction between tokens at each layer. As a result, when processing video data arranged along the temporal axis, the computational cost increases significantly. To mitigate this issue, directly applying the proposed 3D Spatio-Temporal Embedding Module in this study would significantly increase the computational load of the model. Therefore, in this experiment, we changed the stride on the image axis from the original 1 to 2. For the Cycle-MLP, we used the same 3D frontend as the MLP-Mixer, and in this study, we employed the Cycle-MLP-B2 model size.

For the Transformer family's Swin Transformer, we used the smallest model size, Swin-T. For lip reading image patch division, we applied the same 3D Spatio-Temporal Embedding Module, initial patch size, and window size used in SwinLip, with channel dimensions and the number of heads identical to those in the original configuration of Swin-T. This allowed for a fair performance comparison, as the 3D CNN layer for processing lip reading data was applied identically to both the original Swin Transformer and the SwinLip during the experiments. For DAT replacing the last two stages of the Swin Transformer with deformable blocks, we used the same model configuration as Swin-T. Additionally, ConvNeXt was also constructed using the same configuration as Swin Transformer, hence we adopted the corresponding ConvNeXt-T model size. The settings for other experimental strategies were the same as those used in the experiments with the LRW dataset.

Table~\ref{tbl-sup} shows the lip reading performance of the proposed SwinLip and other vision models. It was observed that models from the MLP family generally had less computational load than the baseline. Here, MLP-Mixer was composed of fewer parameters than the baseline but showed 0.42\% lower recognition performance. Cycle-MLP performed better than the MLP-Mixer but had a higher computational load. This indicated that capturing both local and global patterns in lip reading images with Cycle Fully-Connected layers yielded better performance than processing patches with simple MLP functions. However, our SwinLip achieved higher performance with even less computational load. It suggested that models based on self-attention were more suitable for speech recognition tasks like lip reading than MLP computational methods.

When ViT family models were applied to lip reading, all three models showed similar computational loads and had fewer FLOPs compared to the baseline, but they required more parameters. However, the lip reading performance was highest with the Swin Transformer and lowest with DAT. This suggested that the additional offset network operations in the DAT structure did not perform well with relatively small images like lip reading data. The ConvNeXt, which applies an architecture inspired by the Swin Transformer to convolution, still underperformed compared to the Swin Transformer in lip reading. This indicated that the hierarchical self-attention of the Swin Transformer effectively modeled complex patterns and relationships in lip reading data, whereas the ConvNeXt, solely based on convolution operations, lacked the capability to grasp complex patterns in lip reading.

Finally, we compared the performance of the original Swin Transformer and our SwinLip for lip reading. When configuring SwinLip, we considered the following aspects: Instead of simply using the original Swin Transformer as is, we aimed to create a model size optimized for lip reading. Additionally, we integrated the Conformer block not just as an attachment to the model, but within the four-stage hierarchical structure of the Swin Transformer, and for this purpose, we removed BN from the Conformer structure. 
BN was removed from the Conformer structure because it can introduce undesirable effects when applied to smaller batch sizes or sequences of varying lengths. Specifically, when BN is applied to randomly batched word utterance data, it performs normalization independently across each batch, which can hinder the optimization of lip reading models that deal with varying data scales. In our hierarchical design, where the Conformer is tightly integrated into the Swin Transformer’s four-stage structure, the normalization mechanism can lead to instability, particularly when handling non-uniform input sizes or dynamic features. Additionally, removing BN simplifies the model’s architecture, reducing the computational overhead, which can be beneficial for real-time applications or environments with limited computational resources.
As a result, our SwinLip significantly reduced the computational load while enhancing performance compared to the original Swin Transformer. By eliminating redundant computations in the last layer and integrating the proposed Conformer block, we were able to reduce the number of training parameters by approximately 35\% and reduce the inference computational cost by around 30\%, delivering better performance than the baseline~\cite{ma2022training}.

\begin{table}[!t]
\centering
\small
\addtolength{\tabcolsep}{-3pt}
\renewcommand\arraystretch{1.3}
\begin{tabular}{lcccc}
\Xhline{3\arrayrulewidth}
 \multicolumn{1}{c}{\begin{tabular}[c]{@{}c@{}}\textbf{Visual}\\ \textbf{Encoder}\end{tabular}}  & \textbf{FLOPs (G)} & \textbf{Params (M)} &  \textbf{Acc. (\%)} \\ 
\hline\hline
\begin{tabular}[c]{@{}c@{}}ResNet18 (Baseline)~\cite{ma2022training}\end{tabular}  & 10.67        & 52.55         & 89.52        \\ \hline
\begin{tabular}[c]{@{}c@{}}MLP-Mixer~\cite{tolstikhin2021mlp}\end{tabular}  & 4.35       & \textbf{49.25}  & 89.10     \\
\begin{tabular}[c]{@{}c@{}}Cycle-MLP~\cite{chen2022cyclemlp}\end{tabular}  & 6.96       & 71.43          & 90.25     \\ \hline
\begin{tabular}[c]{@{}c@{}}Swin Transformer~\cite{liu2021swin}\end{tabular} & 5.53         & 88.04          & 90.56  \\
\begin{tabular}[c]{@{}c@{}}DAT~\cite{xia2022vision}\end{tabular} & 5.54         & 88.05                 & 88.32       \\
\begin{tabular}[c]{@{}c@{}}ConvNeXt~\cite{liu2022convnet}\end{tabular} & 5.59         & 88.08       & 89.85     \\ \hline
\textbf{SwinLip (Ours)}      & 3.40   & 53.84             & \textbf{90.67}           \\ 
\textbf{SwinLip-Streaming (Ours)}      & \textbf{3.32}   & 51.21        & 90.30           \\\Xhline{3\arrayrulewidth}
\end{tabular}
\caption{Comparison of the FLOPs, numbers of parameters, and word accuracies of our SwinLip and the ResNet18-based baseline, and other vision network models on the LRW dataset. In common, DC-TCN was used as the backend.}\label{tbl-sup}
\end{table}

\subsection{Comparison with State-of-the-Art Models}
We compared the results of the lip reading model applying the proposed SwinLip with word-level lip reading benchmarks for two languages. In this paper, the focus is on replacing with SwinLip the ResNet18-based visual speech encoder, which is widely used in most lip reading studies. We compared the performance of models using only visual information with those utilizing auxiliary techniques. The auxiliary techniques include using audio data as an additional modality or applying KD.

Our SwinLip, designed to efficiently process global visual information, demonstrated superior performance compared to ResNet18, which has established itself as the standard frontend for lip reading models. Table~\ref{tbl5} shows word accuracies on the English LRW dataset, where our SwinLip replacing the ResNet18-based visual speech encoder achieved the word accuracy of 90.67\%, the best accuracy obtained through a VO model. Notably, as shown in Table~\ref{tbl4}, SwinLip reduced the computational load by a third compared to the recent baseline model~\cite{ma2022training}, enabling faster visual feature encoding. In the word boundary mode, which includes the timing information of word utterances, SwinLip achieved the word accuracy of 92.43\%. This performance was a similar accuracy with a lower computational cost than that of the method that inserted the CRO-TSM module into the ResNet18 architecture~\cite{xiang2024collaboration}. 

Most current high-performance lip reading methods rely on audio information to compensate for the limitations of visual features. For example, SyncVSR~\cite{ahn2024syncvsr} proposed a training technique that synchronizes audio tokens with visual features, which allowed for maintaining high performance even with limited data by quantizing audio tokens to better align with visual information. On the other hand, MVM~\cite{kim2022distinguishing}, and MTLAM~\cite{yeo2023multi} stored both visual and audio features in memory to address the challenge of homophenes, enhancing the mapping between visual articulation and corresponding speech. While these methods effectively supplemented the lack of visual information, they might heavily depend on audio data.
In the KD method, WPCL+APFF~\cite{tian2022lipreading} applied two identical lip reading models, adding post-processing modules that capture partial lip regions for KD. Additionally, \cite{ma2021towards, ma2022training} applied a self-distillation method using five identical models. While these KD methods can achieve better performance, they also result in significant computational overhead.
\begin{table}[!t]
\centering
\small
\addtolength{\tabcolsep}{-1.pt}
\renewcommand\arraystretch{1.3}
\begin{tabular}{lccc}
\Xhline{3\arrayrulewidth}
\multicolumn{1}{c}{\textbf{Method (without Word Boundary)}}                          & \textbf{Acc. (\%)}    \\ 
\hline\hline
VGG-M + LSTM~\cite{chung2017lip}                                      & 61.1                 \\ 
ResNet34 + Bi-LSTM~\cite{stafylakis2017combining}            & 83.0                 \\ 
2$\times$ResNet18 + Bi-GRU (DFTN)~\cite{xiao2020deformation} & ~~84.13                \\
ShuffleNetV2 + MS-TCN~\cite{ma2021towards}                   & 84.4                 \\
ResNet18 + Bi-GRU~\cite{feng2020learn}                       & 85.0                 \\
ResNet18 + MS-TCN~\cite{martinez2020lipreading}              & 85.3                 \\
ResNet18 + Bi-GRU + TSM~\cite{hao2021use}                    & ~~86.23                \\
ResNet18 + Bi-GRU + CRO-TSM~\cite{xiang2024collaboration}    & 88.9                 \\
EfficientNetV2-L + TCN + Transformer~\cite{koumparoulis2022accurate}        & ~~89.52                \\
ResNet18 + DC-TCN~\cite{ma2022training}                      & 90.4                 \\ \hline
\textbf{SwinLip + DC-TCN (Ours)}                                    & \textbf{~~90.67}       \\ \Xhline{3\arrayrulewidth}
\multicolumn{1}{c}{\textbf{Method (with Word Boundary)}}     & \textbf{Acc. (\%)}    \\ \hline\hline
ResNet18 + Bi-GRU + WB~\cite{feng2020learn}                  & 88.4                 \\
3DCvT + Bi-GRU + WB~\cite{wang2022lip}                       & 88.5                 \\
ResNet18 + DC-TCN + WB~\cite{ma2022training}                 & 92.1                 \\ 
ResNet18 + DC-TCN + WB + CRO-TSM~\cite{xiang2024collaboration}    & 92.4                 \\\hline
\textbf{SwinLip + DC-TCN + WB (Ours)}                               & \textbf{~~92.43}       \\ \Xhline{3\arrayrulewidth}
\multicolumn{1}{c}{\textbf{Method (with Auxiliary Techniques)}}     & \textbf{Acc. (\%)}    \\ \hline\hline
ResNet18 + MS-TCN + KD (Ensemble)~\cite{ma2021towards}       & 88.5                 \\
ResNet18 + MS-TCN + MVM~\cite{kim2022distinguishing}    & ~88.5$^{*}$  \\
ResNet18 + MS-TCN + WPCL + APFF~\cite{tian2022lipreading}    & 88.3          \\
ResNet18 + DC-TCN + WB + KD (Ensemble)~\cite{ma2022training}   & 94.1 \\ 
ResNet18 + DC-TCN + MTLAM~\cite{yeo2023multi}   & ~91.7$^{*}$  \\
ResNet18 + Transformer + SyncVSR~\cite{ahn2024syncvsr} & ~93.2$^{*}$ \\
ResNet18 + Transformer + WB + SyncVSR~\cite{ahn2024syncvsr} & \textbf{~95.0$^{*}$} \\ \Xhline{3\arrayrulewidth}
\end{tabular}
\caption{Accuracies of the presented SwinLip and state-of-the-art works of lip reading on the LRW dataset. Auxiliary techniques refer to methods where audio modality-based ASR encoders or knowledge distillation (KD) were applied. $^{*}$ indicates that audio data was used.}\label{tbl5}
\end{table}

\begin{table}[!t]
\centering
\small
\addtolength{\tabcolsep}{6.pt}
\renewcommand\arraystretch{1.3}
\begin{tabular}{lccc}
\Xhline{3\arrayrulewidth}
\multicolumn{1}{c}{\textbf{Method (without Word Boundary)}}                          & \textbf{Acc. (\%)}   \\  \hline\hline
ResNet34 + Bi-LSTM~\cite{yang2019lrw}                        & ~~38.19               \\ 
ResNet18 + MS-TCN~\cite{martinez2020lipreading}              & 41.4                \\
2$\times$ResNet18 + Bi-GRU (DFTN)~\cite{xiao2020deformation} & ~~41.93               \\
ResNet18 + DC-TCN~\cite{ma2021lip}                           & ~~43.65               \\
ResNet18 + Bi-GRU + TSM~\cite{hao2021use}                    & ~~44.60               \\
ResNet18 + Bi-GRU~\cite{feng2020learn}                       & 48.0                \\ \hline 
\textbf{SwinLip + Bi-GRU (Ours)}                        & \textbf{~~48.09}      \\ \Xhline{3\arrayrulewidth}
\multicolumn{1}{c}{\textbf{Method (with Word Boundary)}}     & \textbf{Acc. (\%)}    \\ \hline\hline
ResNet18 + Bi-GRU + WB~\cite{feng2020learn}                  & 55.7                \\
3DCvT + Bi-GRU + WB~\cite{wang2022lip}                       & 57.5                \\  \hline
\textbf{SwinLip + Bi-GRU + WB (Ours)}                        & \textbf{~~59.41}      \\ \Xhline{3\arrayrulewidth}
\multicolumn{1}{c}{\textbf{Method (with Auxiliary Techniques)}}     & \textbf{Acc. (\%)}    \\ \hline\hline
ResNet18 + MS-TCN + KD (Ensemble)~\cite{ma2021towards}       & 46.6                \\
ResNet18 + MS-TCN + WPCL + APFF~\cite{tian2022lipreading}    & 49.4                \\
ResNet18 + MS-TCN + MVM~\cite{kim2022distinguishing}    & ~53.8$^{*}$  \\
ResNet18 + DC-TCN + MTLAM~\cite{yeo2023multi}   & ~54.3$^{*}$  \\
ResNet18 + Transformer + SyncVSR~\cite{ahn2024syncvsr} & \textbf{~58.2$^{*}$} \\ \Xhline{3\arrayrulewidth}
\end{tabular}
\caption{Accuracies of the presented SwinLip and state-of-the-art studies of lip reading on the LRW-1000 dataset. Auxiliary techniques refer to methods where audio modality-based ASR encoders or knowledge distillation (KD) were applied. $^{*}$ indicates that audio data was used.}\label{tbl6}
\end{table}

Our SwinLip achieved high performance using only visual information, without relying on such audio-based auxiliary techniques. SwinLip can be seamlessly integrated into various backend architectures of lip reading, optimizing lip reading performance without needing audio information. This is useful for handling language differences or complex phonetic structures. We evaluated SwinLip on the Mandarin LRW-1000 dataset, which is relatively challenging due to noise levels and a large number of classes. Table~\ref{tbl6} shows a performance comparison with previous state-of-the-art works on the LRW-1000 dataset. Our SwinLip achieved the word accuracy of 59.41\%, surpassing all previous works. This result demonstrates that our SwinLip, even without audio data, achieved state-of-the-art performance in complex languages like Mandarin, showing great potential for various VSR tasks.

While auxiliary techniques and modules such as TSM increase the overall computational load, our SwinLip delivered comparable or better performance without these additional computational costs. These results showed the strong potential of SwinLip to replace ResNet18 in various lip reading tasks.

\section{Conclusion}
In this paper, we presented SwinLip, an efficient visual speech encoder for lip reading, which featured a new lightweight scale of the Swin Transformer and incorporated temporal embeddings into its hierarchical structure. Experimental results showed that our SwinLip significantly improved recognition performance while substantially reducing computational load on both word- and sentence-level tasks by seamlessly combined with various speech recognition backends. Our SwinLip showed robust performance across different languages, including English and Mandarin, and notably achieved new state-of-the-art performance on the LRW-1000 benchmark dataset while reducing computational load. This demonstrated that the proposed SwinLip could be efficiently used as a visual speech encoder for lip reading. Since our SwinLip has been confirmed to be compatible with various backends, it is expected to be efficiently integrated with KD and audio-augmented models, and we plan to conduct research on this integration in the future.

\section*{Acknowledgement}
This work was partly supported by the National Research Foundation of Korea (NRF) and the Commercialization Promotion Agency for R\&D Outcomes (COMPA) grant funded by the Korea government (MSIT) (RS-2023-00237117) and Institute of Information \& communications Technology Planning \& Evaluation (IITP) grant funded by the Korea government (MSIT) (No. 2022-0-00621, RS-2022-II220621, Development of artificial intelligence technology that provides dialog-based multi-modal explainability).

\bibliographystyle{elsarticle-num} 
\bibliography{ref}





\end{document}